\pdfoutput=1

\documentclass[11pt]{article}

\usepackage[]{acl}

\usepackage{times}
\usepackage{latexsym}

\usepackage[T1]{fontenc}

\usepackage[utf8]{inputenc}

\usepackage{microtype}

\usepackage{graphicx}

\usepackage{inconsolata}

\usepackage{booktabs} 
\usepackage{multirow}
\usepackage{bm}
\usepackage{array}
\usepackage[shortlabels]{enumitem}
\usepackage{tikz}
\usepackage{pgfplots}
\usepackage[hypcap=true]{subcaption}
\usepackage{arydshln}
\usepackage{makecell}

\title{Improving the Downstream Performance of Mixture-of-Experts Transformers via Weak Vanilla Transformers}

\author{Xin Lu,\ \ Yanyan Zhao\thanks{\ \ \  Email corresponding.},\ \ Bing Qin,\ \ Ting Liu \\
	Research Center for Social Computing and Interactive Robotics\\
	Harbin Institute of Technology, China\\
	\texttt{\{xlu, yyzhao, qinb, tliu\}@ir.hit.edu.cn}
}

\begin{document}
\maketitle
\begin{abstract}
Recently, Mixture of Experts (MoE) Transformers have garnered increasing attention due to their advantages in model capacity and computational efficiency. 
However, studies have indicated that MoE Transformers underperform vanilla Transformers in many downstream tasks, significantly diminishing the practical value of MoE models. 
To explain this issue, we propose that the pre-training performance and transfer capability of a model are joint determinants of its downstream task performance. 
MoE models, in comparison to vanilla models, have poorer transfer capability, leading to their subpar performance in downstream tasks. 
To address this issue, we introduce the concept of transfer capability distillation, positing that although vanilla models have weaker performance, they are effective teachers of transfer capability. 
The MoE models guided by vanilla models can achieve both strong pre-training performance and transfer capability, ultimately enhancing their performance in downstream tasks. 
We design a specific distillation method and conduct experiments on the BERT architecture. 
Experimental results show a significant improvement in downstream performance of MoE models, and many further evidences also strongly support the concept of transfer capability distillation. 
Finally, we attempt to interpret transfer capability distillation and provide some insights from the perspective of model feature. 
\end{abstract}

\begin{figure}[!t]
	\setlength{\belowcaptionskip}{-0.09cm}
	\centering
	\includegraphics[scale=0.725]{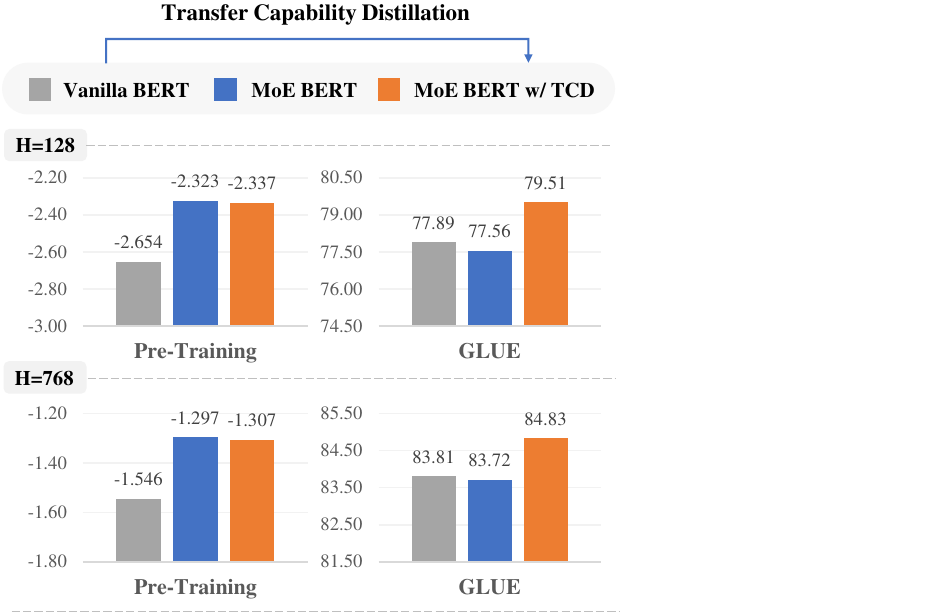}
	\caption{ The results of pre-training and fine-tuning show that: 1) The original MoE models have poor downstream performance, which is related to inferior transfer capability; 2) The MoE models with transfer capability distillation exhibit significant improvement; 3) The teacher models have weaker performance, hence this distillation only involves their strong transfer capability. }
	\label{figure_1}
\end{figure}

\section{Introduction}

Recent researches have revealed that pre-trained language models demonstrate powerful general capabilities~\cite{devlin-etal-2019-bert,NEURIPS2020_1457c0d6,NEURIPS2022_b1efde53,openai2023gpt4} and an extraordinary ability to enhance performance through scaling~\cite{kaplan2020scaling,NEURIPS2022_c1e2faff}. 
However, scaling up these models incurs significant costs in practical applications due to the rapid increase in computational demands. 
As a result, there is a growing interest in Mixture of Experts (MoE) models~\cite{6797059,shazeer2017,lepikhin2021gshard,pmlr-v162-du22c}. 
These models enable inputs to be processed by distinct experts. 
The number of experts determines the number of parameters while having a limited effect on computation cost, thereby expanding the capacity with lower computation expense.

However, existing researches indicate that while MoE models excel in pre-trained language modeling tasks, their efficacy diminishes in downstream tasks, especially when a large number of experts are involved. 
\citet{JMLR_v23_21_0998} proposed the Switch Transformers based on MoE architecture, revealed that MoE models consistently underperform vanilla models in fine-tuning on SuperGLUE benchmark~\cite{NEURIPS2019_4496bf24} when pre-training performances are equivalent. 
\citet{artetxe-etal-2022-efficient} conducted more experiments, and it can also be observed from their published results that MoE models consistently show weaker fine-tuning results in downstream tasks when pre-training performances are equivalent. 
\citet{shen2023mixtureofexperts} similarly observed in their experiments that, on many downstream tasks, single-task fine-tuned MoE models underperform their dense counterparts.

\begin{figure}[!t]
	\setlength{\belowcaptionskip}{-0.20cm}
	\centering
	\includegraphics[scale=0.55]{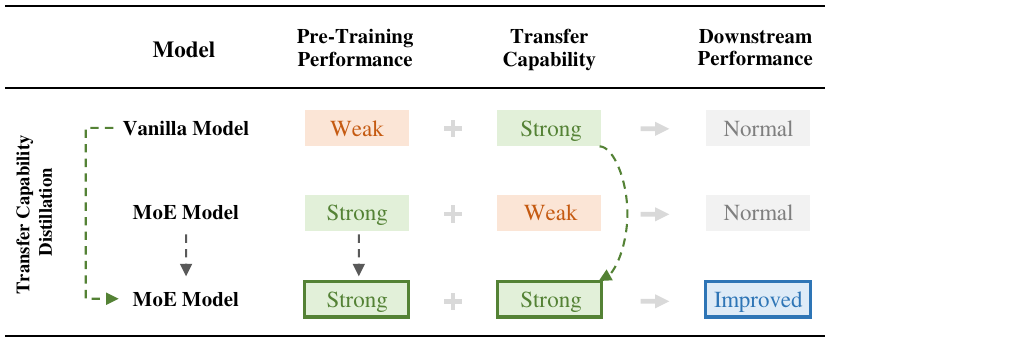}
	\caption{ The mechanism diagram of transfer capability distillation. It enhances the transfer capability of MoE models. Combined with their strong pre-training performance, the downstream performance can be improved. }
	\label{figure_2}
\end{figure}

We also conduct relevant validation, pre-training two scales of vanilla BERT models and MoE BERT models with 64 experts (top-1 activation), followed by fine-tuning on GLUE benchmark~\cite{wang-etal-2018-glue}. 
Some results are shown in Figure \ref{figure_1}. 
We observe that for both scales, the MoE models need to improve pre-training performance (Log likelihood) to a much higher level to achieve similar GLUE performance to the vanilla models. 
This implies that the pre-training performance gains brought about by introducing multiple experts in MoE models do not translate effectively to the performance improvement of downstream tasks we are primarily concerned with, thereby significantly diminishing the practical value of MoE models. 

We attempt to address this issue. 
Initially, we need to explain the poor performance in downstream tasks of the MoE models. 
\textbf{We believe the downstream performance of a model is determined by both pre-training performance and transfer capability.} 
Pre-training performance is acquired through training, whereas transfer capability is an inherent attribute of the model. 
The latter determines the extent to which the former can be converted into downstream performance. 
Vanilla models, despite their smaller capacity and weaker pre-training performance, possess strong transfer capability. 
In contrast, MoE models, although having larger capacity and stronger pre-training performance, only exhibit weak transfer capability. 
Therefore, \textbf{we believe the poor performance of the MoE models in downstream tasks is primarily due to its limited transfer capability}, as summarized in Figure \ref{figure_2}. 

Based on the above explanation, we propose a solution to this issue: 
since the transfer capability of vanilla models is strong, is it possible to transfer this capability to MoE models through distillation? 
\textbf{We call this idea Transfer Capability Distillation (TCD). }
The underlying logic is that although the pre-training and downstream performance of vanilla models are relatively weak, the transfer capability of them is stronger. 
By using them as teachers, we might enhance the transfer capability of MoE models. 
Combined with the strong pre-training performance of MoE models, this approach could lead to a comprehensive improvement of MoE models, as depicted in Figure \ref{figure_2}.

The most counterintuitive feature of this method is: 
\textbf{a teacher model, inferior in pre-training and downstream performance, anomalously distills a student model superior in those aspects}. 

Based on the above ideas, we design a distillation scheme and conduct experiments. 
Some results are shown in Figure \ref{figure_1}. 
The results indicate the downstream performance of the MoE model with TCD, not only improved over the original MoE model but also exceeded that of its teacher model. 
This supports the concept of transfer capability distillation, successfully improving the MoE models. 

Moreover, we also conduct a more discussion, providing insights into the differences in transfer capability from model feature perspective, and explaining why our distillation can be effective. 

The contributions of our work are as follows:

\begin{itemize}[topsep=1.5pt,itemsep=1.5pt]
	\item  We differentiate pre-training performance from transfer capability as distinct influencers of downstream performance, identifying the cause of poor downstream performance in MoE models as inferior transfer capability. 
	\item  We introduce transfer capability distillation, identifying vanilla transformers as effective teachers and proposing a distillation scheme. 
	\item  By transfer capability distillation, we address the issue of weak transfer capability in MoE models, enhancing downstream performance. 
	\item  We provide insights into the differences in transfer capability from model feature perspective and offer a basic explanation of the mechanisms of transfer capability distillation. 
\end{itemize}

\section{Method}

\subsection{Overview}

In this work, we propose a transfer capability distillation scheme. 
The core idea is as follows: 

First, a teacher model with low capacity but strong transfer capability is pre-trained, which exhibits weaker performance in both pre-training and downstream tasks. 
Then, during the pre-training of high-capacity student model, not only original pre-training loss is optimized, but a new transfer capability distillation loss is also optimized. 
Finally, the student model acquires strong transfer capability on top of strong pre-training performance,  achieving transfer capability distillation. 

In following sections, we will first introduce the vanilla BERT model as the teacher model and the MoE BERT as the student model. 
Subsequently, we will introduce the specific implementation of transfer capability distillation, and conclude with an overview of the training process.

\subsection{Vanilla BERT and MoE BERT}

Our work concerns two BERT architectures within Transformers: Vanilla BERT and MoE BERT. 
The vanilla BERT has a smaller capacity and weaker pre-training performance but exhibits strong transfer capability, making it suitable as a teacher model. 
The MoE BERT has a larger capacity and stronger pre-training performance but weaker transfer capability, serving as the student model. 

\begin{figure}[!t]
	\setlength{\belowcaptionskip}{-0.03cm}
	\centering
	\includegraphics[scale=0.38]{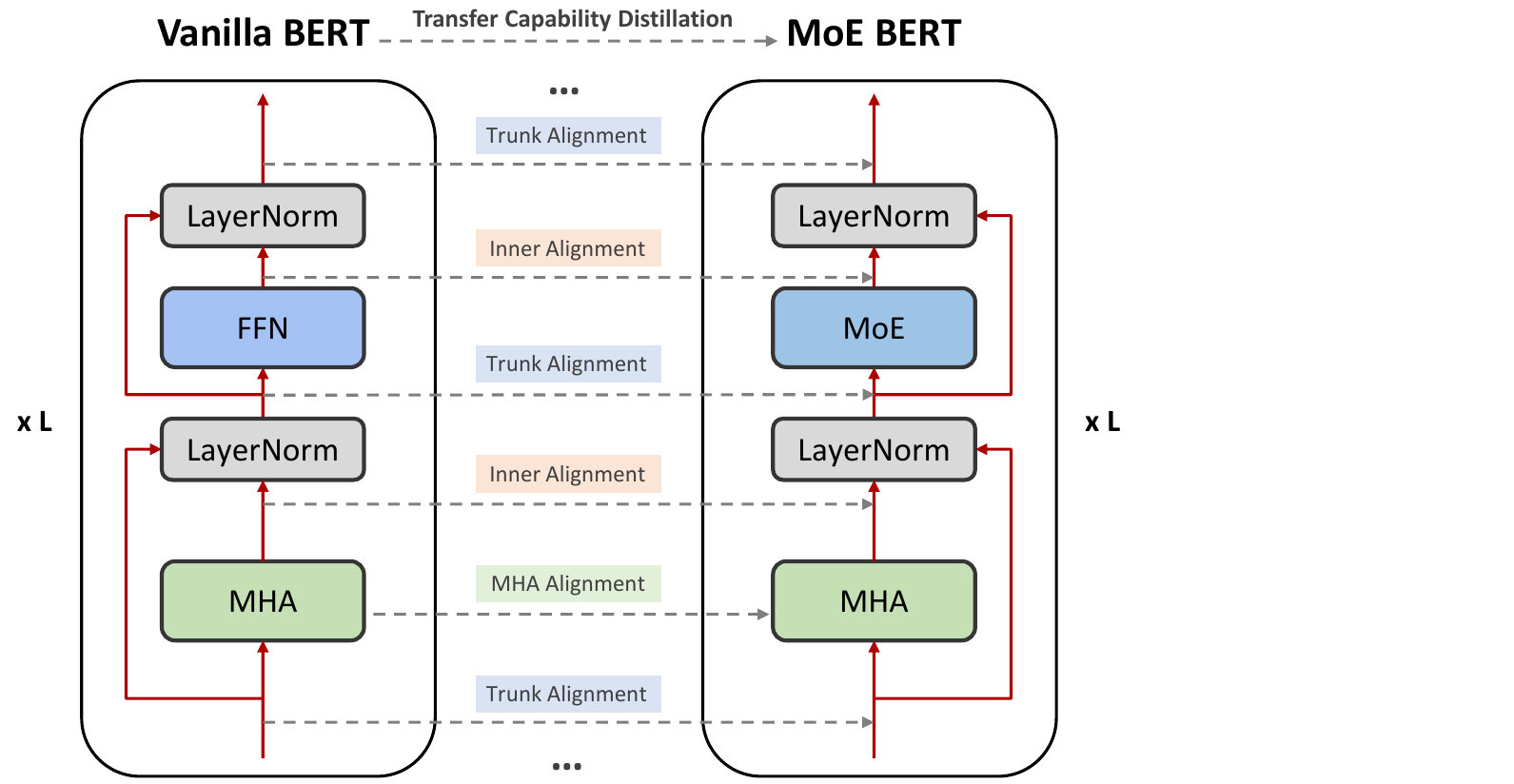}
	\caption{ Overview of our transfer capability distillation scheme. It involves relation alignment in three locations. }
	\label{figure_3}
\end{figure}

The structure of the vanilla BERT, as shown on the left side of Figure \ref{figure_3}, consists of stacked Multi-Head Attention (MHA) and Feed-Forward Networks (FFN), employing a post layer normalization scheme for residuals and normalization. 
We follow the structure design by \citet{devlin-etal-2019-bert}, retaining the original structure of the BERT model. 
We denote the original masked language modeling loss in pre-training phase as $\mathcal{L}_{MLM}$. 

\begin{figure}[!t]
	\centering
	\includegraphics[scale=0.417]{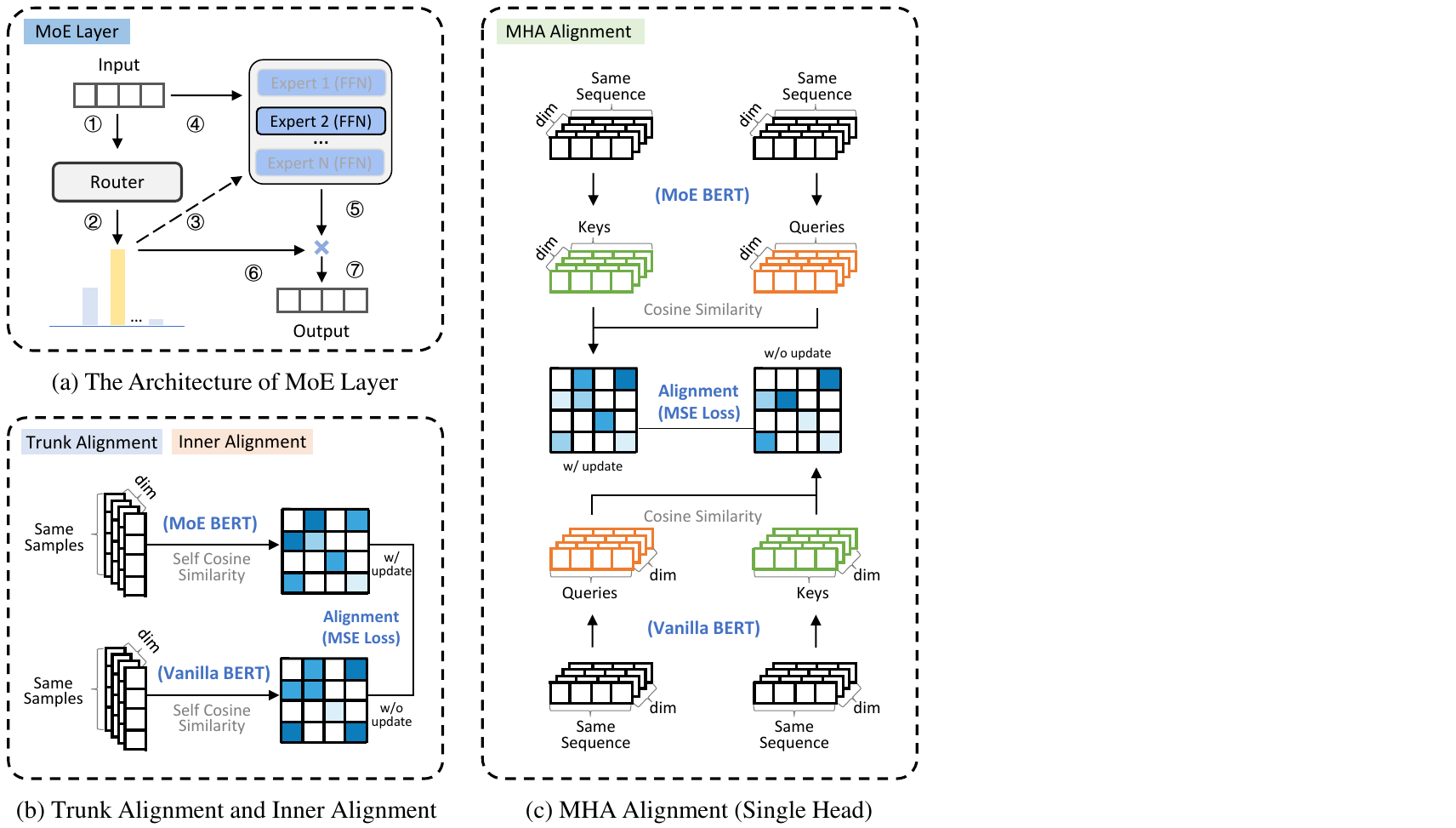}
	\caption{ Details of our proposed distillation scheme. }
	\label{figure_4}
\end{figure}

The structure of the MoE BERT, as shown on the right side of Figure \ref{figure_3}, differs from the vanilla BERT by replacing all FFN layers with MoE layers. 
The basic structure of an MoE layer, as illustrated in Figure \ref{figure_4}(a), is not composed of a single FFN but includes multiple FFNs, also known as multiple experts. 
When the hidden representation of a token is fed into an MoE layer, a routing module (linear layer with softmax) first predicts the probability of it being processed by each expert, and then the hidden representation of the token is only processed by the top-k experts in terms of probability.

Assume the hidden representation is $\mathbf x$, and the parameters of the routing module are $\mathbf{W}_{r}$ and $\mathbf{b}_{r}$, then the process of calculating the probability of selecting each expert is as follows: 

\begin{equation}
	p(\mathbf x) = {\rm softmax}(\mathbf{W}_{r}\mathbf{x} + \mathbf{b}_{r})
\end{equation}

In this work, we adhere to two key practices of the Switch Transformers~\cite{JMLR_v23_21_0998}: 

1. Only the top-1 expert in terms of probability processes the hidden representation. 
The process for determining the expert index is as follows: 

\begin{equation}
	i=\mathop{\rm argmax}\limits_{k}{\ p_k(\mathbf{x})}
\end{equation}

2. The hidden representation of the token is first processed by the expert, and then multiplied by the probability of selecting that expert to obtain the final representation. 
This strategy enables effective gradient descent optimization for the routing module. 
Assume the set of all experts is $ \{{E_k(\mathbf{x})}\}_{k=1}^N $, and the processing is as follows: 

\begin{equation}
	\mathbf{h}=p_i(\mathbf{x})E_i(\mathbf{x})
\end{equation}

Additionally, for expert load balancing, we calculate the Kullback-Leibler divergence between the average probability distribution of experts selected within a batch and a uniform distribution, adding it as an additional loss term. 

Assuming there are $ M $ hidden representations in a batch and the vector of uniform probability distribution is $ \mathbf{p} $, then this process is as follows: 

\begin{equation}
	\mathbf{q}=\frac{1}{M}\sum_{j=1}^{M}{softmax(\mathbf{W}_r\mathbf{x}_{j} + \mathbf{b}_{r})}
\end{equation}

\begin{equation}
	\mathcal{L}_B=KL(\mathbf{p}||\mathbf{q})
\end{equation}

\subsection{Transfer Capability Distillation}

Although the transfer capability distillation in this work differs in background and final influence from general knowledge distillation, the implementation strategy is similar, namely, it is achieved by aligning the representations in the intermediate layers between the student and teacher models. 

Different from existing works~\cite{sun-etal-2019-patient,sanh2020distilbert,jiao-etal-2020-tinybert,sun-etal-2020-mobilebert}, we avoid direct alignment of intermediate layer representations, i.e., we do not use Mean Squared Error (MSE) to make the values of individual sampled representations converge; 
instead, we choose to align the relationships between representations, that is, to make the cosine similarity of a pair of sampled representations converge. 

We consider that direct alignment imposes too strict limitations on the values of representations. 
Since the teacher model is a pre-trained model with weaker performance, in extreme cases, this could lead to a complete degradation of the student model's pre-training performance to the level of the teacher model, rendering the transfer capability distillation meaningless. 
By opting to align the relationships between representations, more flexibility is provided for the values of representations, potentially reducing conflicts between pre-training objective and distillation objective. 
In our experiments, we indeed found that this approach results in transfer capability distillation that does not compromise pre-training performance.

\linespread{1.3}
\begin{table*}[t]
	\linespread{1.0}
	\caption{ Experimental results of Pre-training Performance Alignment settings on the dev set of GLUE Benchmark. }
	\linespread{1.3}
	\label{table_1}
	\resizebox{1.0\textwidth}{!}{
		\begin{tabular}{@{}clcccccccccccccccc@{}}
			\toprule
			& \multirow{2}{*}{\textbf{Model}}   & \textbf{Pre-Train} & \textbf{Pre-Train} &  \textbf{CoLA} & \multicolumn{2}{c}{\textbf{MRPC}} & \textbf{SST-2} & \multicolumn{2}{c}{\textbf{STS-B}} & \textbf{RTE} & \multicolumn{2}{c}{\textbf{MNLI}} & \textbf{QNLI} & \multicolumn{2}{c}{\textbf{QQP}} & \textbf{Avg.} & \\ 
			&        & \textbf{Epoch} & \textbf{Pref.} & (8.5k) & \multicolumn{2}{c}{(3.7k)} & (67k) & \multicolumn{2}{c}{(7.0k)} & (2.5k) & \multicolumn{2}{c}{(393k)} & (108k) & \multicolumn{2}{c}{(364k)} & \textbf{Score} & \\ 
			\midrule
			\specialrule{0em}{0pt}{4pt}
			\multicolumn{5}{l}{\textit{Pre-Training Performance Alignment (\textbf{H=128})}} & & & & & & & & & & & & & \\
			\specialrule{0em}{0pt}{4pt}
			& \textit{Vanilla BERT (Teacher)} & 20.0 & \ -2.65387 \  & 33.88  & 83.03 & 88.09 & 86.65 & 83.44 & 83.39 & 63.41 & 76.77 & 77.36 & 85.15 & 88.71 & 84.82 & 77.89  & \\ 
			\specialrule{0em}{0pt}{4pt}
			& MoE BERT                        &  6.0 & \ -2.32278 \  & 37.56 & 82.11 & 87.15 & 86.22 & 83.68 & 83.31 & 62.86 & 74.94 & 76.13 & 84.97 & 87.87 & 83.90 & 77.56  & \\ 
			\specialrule{0em}{0pt}{4pt}
			& MoE BERT w/ TCD                 &  5.0 & \ -2.33650 \  & \textbf{44.10} & \textbf{84.52} & \textbf{89.09} & \textbf{87.41} & \textbf{84.08} & \textbf{83.81} & \textbf{65.70} & \textbf{77.27} & \textbf{78.29} & \textbf{86.12} & \textbf{88.72} & \textbf{85.01} & \textbf{79.51}  & \\ 
			\specialrule{0em}{0pt}{4pt}
			\midrule
			\specialrule{0em}{0pt}{4pt}
			\multicolumn{5}{l}{\textit{Pre-Training Performance Alignment (\textbf{H=768})}} & & & & & & & & & & & & & \\
			\specialrule{0em}{0pt}{4pt}
			& \textit{Vanilla BERT (Teacher)} & 10.0 & \ -1.54597 \  & 62.03 & 86.93 & 90.65 & 92.73 & 87.64 & 87.31 & 61.97 & 83.73 & 83.88 & 90.72 & 90.71 & 87.44 & 83.81  & \\
			\specialrule{0em}{0pt}{4pt} 
			& MoE BERT                        & 12.0 & \ -1.29679 \  & 64.24 & 86.11 & 90.29 & 93.12 & 87.25 & 86.89 & 61.73 & 83.61 & 83.61 & 90.26 & 90.29 & 87.26 & 83.72  & \\ 
			\specialrule{0em}{0pt}{4pt}
			& MoE BERT w/ TCD                 & 10.0 & \ -1.30669 \  & \textbf{65.36} & \textbf{88.03} & \textbf{91.53} & \textbf{93.46} & \textbf{88.10} & \textbf{87.79} & \textbf{64.14} & \textbf{84.65} & \textbf{84.68} & \textbf{91.63} & \textbf{90.85} & \textbf{87.70} & \textbf{84.83}  & \\ 
			\specialrule{0em}{0pt}{4pt}
			\bottomrule
		\end{tabular}
	}
\end{table*}
\linespread{1.0}

Specifically, we select three locations in the vanilla BERT models and MoE BERT models for relation alignment, as shown in Figure \ref{figure_3}.

\textbf{Model Trunk} \ \ After the layer normalization in all MHA layers and FFN layers, we add relation constraints to the normalized hidden representations. 
Specifically, multiple tokens are randomly selected from a batch, and for any pair of tokens, the cosine similarity of their normalized hidden representations is calculated. 
The similarity computed by the student model is aligned with that computed by the teacher model, as shown in Figure \ref{figure_4}(b). 

Suppose the set of tokens selected from a batch is $ \{{x_k}\}_{k=1}^N $, the student model's normalized hidden representations are $ \{{\mathbf{h}_k}\}_{k=1}^N $, and the teacher model's normalized hidden representations are $ \{{\mathbf{h}_k^\prime}\}_{k=1}^N $; then, the process is as follows: 
\begin{equation}
	s_{ij} = \frac{\mathbf{h}_i \cdot \mathbf{h}_j}{\|\mathbf{h}_i\| \|\mathbf{h}_j\|}
\end{equation}

\begin{equation}
	s_{ij}^\prime = \frac{\mathbf{h}_i^\prime \cdot \mathbf{h}_j^\prime}{\|\mathbf{h}_i^\prime\| \|\mathbf{h}_j^\prime\|}
\end{equation}

\begin{small}
\begin{equation}
	\mathcal{L}_{Trunk}^\ast=\frac{1}{N^2}\sum_{i=1}^{N}\sum_{j=1}^{N}{{\rm MSE}(s_{ij},s_{ij}^\prime)}
\end{equation}
\end{small}

\textbf{Residual Inner} \ \ Before the layer normalization in all MHA layers and FFN layers, we add relation constraints to the hidden representations that have not undergone residual connections. 
It is similar to that in the model trunk, as detailed in Figure \ref{figure_4}(b). The loss calculated is denoted as $ \mathcal{L}_{Inner}^\ast $. 

\textbf{Multi-Head Attention} \ \ Within all MHA layers, we calculate the cosine similarity between the query and key pairs, aligning the similarity computed by the student model with that computed by the teacher model, as shown in Figure \ref{figure_4}(c). 

For a single head within an MHA layer, the student model's query and key representations are denoted as $ \{{\mathbf{q}_k}\}_{k=1}^M $ and $ \{{\mathbf{k}_k}\}_{k=1}^M $, and the teacher model's as $ \{{\mathbf{q}_{k}^\prime}\}_{k=1}^M $ and $ \{{\mathbf{k}_{k}^\prime}\}_{k=1}^M $, respectively. This process is as follows: 
\begin{equation}
	s_{ij} = \frac{\mathbf{q}_i \cdot \mathbf{k}_j}{\|\mathbf{q}_i\| \|\mathbf{k}_j\|}
\end{equation}

\begin{equation}
	s_{ij}^\prime = \frac{\mathbf{q}_i^\prime \cdot \mathbf{k}_j^\prime}{\|\mathbf{q}_i^\prime\| \|\mathbf{k}_j^\prime\|}
\end{equation}

\begin{small}
\begin{equation}
	\mathcal{L}_{Attention}^\prime=\frac{1}{M^2}\sum_{i=1}^{M}\sum_{j=1}^{M}{{\rm MSE}(s_{ij},s_{ij}^\prime)}
\end{equation}
\end{small}

The loss for a single head is noted as $ \mathcal{L}_{Attention}^\prime $, and the average loss for multiple heads within a batch is noted as $ \mathcal{L}_{Attention}^\ast $. 

The total loss from three constraints is recorded as $ \mathcal{L}_T $, $ \mathcal{L}_I $, and $ \mathcal{L}_A $, corresponding to the total of all positions $ \mathcal{L}_{Trunk}^\ast $, $ \mathcal{L}_{Inner}^\ast $, and $ \mathcal{L}_{Attention}^\ast $.

\subsection{Training Process}

We introduce the main process of training a MoE BERT with transfer capability distillation. 

First, the vanilla BERT is pre-trained to serve as the transfer capability teacher model. 
This model receives original masked language modeling training and achieves baseline performance in both pre-trained and downstream. The pre-training loss of this model is: 

\begin{small}
\begin{equation}
	\mathcal{L}=\mathcal{L}_{MLM}
\end{equation}
\end{small}

Next, the MoE BERT model is pre-trained. 
This model not only optimizes masked language modeling loss and load balancing loss but also uses the vanilla BERT as a transfer capability teacher model, calculating and optimizing multiple distillation losses. 
Its pre-training loss is: 

\begin{small}
\begin{equation}
	\mathcal{L}=\mathcal{L}_{MLM}+\lambda_B\mathcal{L}_B+{\lambda_T\mathcal{L}}_T+\lambda_I\mathcal{L}_I+{\lambda_A\mathcal{L}}_A
\end{equation}
\end{small}

Ultimately, we obtain a MoE BERT enhanced through transfer capability distillation, which has stronger transfer capability compared to an original pre-trained MoE BERT.

\linespread{1.3}
\begin{table*}[t]
	\linespread{1.0}
	\caption{ Experimental results of Pre-training Epoch Alignment settings on the dev set of GLUE Benchmark. \ \ }
	\linespread{1.3}
	\label{table_2}
	\resizebox{1.0\textwidth}{!}{
		\begin{tabular}{@{}clccccccccccccccccc@{}}
			\toprule
			& \multirow{2}{*}{\textbf{Model}}   & \textbf{Teacher} & \textbf{Pre-Train} & \textbf{Pre-Train} &  \textbf{CoLA} & \multicolumn{2}{c}{\textbf{MRPC}} & \textbf{SST-2} & \multicolumn{2}{c}{\textbf{STS-B}} & \textbf{RTE} & \multicolumn{2}{c}{\textbf{MNLI}} & \textbf{QNLI} & \multicolumn{2}{c}{\textbf{QQP}} & \textbf{Avg.} & \\ 
			&        & \textbf{Epoch}& \textbf{Epoch} & \textbf{Pref.} & (8.5k) & \multicolumn{2}{c}{(3.7k)} & (67k) & \multicolumn{2}{c}{(7.0k)} & (2.5k) & \multicolumn{2}{c}{(393k)} & (108k) & \multicolumn{2}{c}{(364k)} & \textbf{Score} & \\ 
			\midrule
			\specialrule{0em}{0pt}{5pt}
			\multicolumn{6}{l}{\textit{Pre-Training Epoch Alignment (\textbf{H=128})}} & & & & & & & & & & & & & \\
			\specialrule{0em}{1.5pt}{1.5pt}
			& MoE BERT                        & - &  25.0 & \ -2.08134 \  & 42.72 & 82.75 & 87.57 & \textbf{87.64} & \textbf{84.10} & \textbf{84.02} & 62.98 & 76.46 & 77.47 & 86.07 & 88.17 & 84.26 & 78.68  & \\ 
			\specialrule{0em}{1.5pt}{1.5pt}
			& MoE BERT w/ TCD                 & 20.0 &  5.0 & \ -2.33650 \  & \textbf{44.10} & \textbf{84.52} & \textbf{89.09} & 87.41 & 84.08 & 83.81 & \textbf{65.70} & \textbf{77.27} & \textbf{78.29} & \textbf{86.12} & \textbf{88.72} & \textbf{85.01} & \textbf{79.51}  & \\ 
			\specialrule{0em}{3pt}{0pt}
			\midrule
			\specialrule{0em}{0pt}{5pt}
			\multicolumn{6}{l}{\textit{Pre-Training Epoch Alignment (\textbf{H=768})}} & & & & & & & & & & & & & \\
			\specialrule{0em}{1.5pt}{1.5pt}
			& MoE BERT                        & - &  20.0 & \ -1.20991 \  & 64.81 & 86.51 & 90.52 & \textbf{93.77} & 87.46 & 87.07 & 62.09 & 84.17 & 84.13 & 90.43 & 90.57 & 87.45 & 84.08  & \\ 
			\specialrule{0em}{1.5pt}{1.5pt}
			& MoE BERT w/ TCD                 & 10.0 &  10.0 & \ -1.30669 \  & \textbf{65.36} & \textbf{88.03} & \textbf{91.53} & 93.46 & \textbf{88.10} & \textbf{87.79} & \textbf{64.14} & \textbf{84.65} & \textbf{84.68} & \textbf{91.63} & \textbf{90.85} & \textbf{87.70} & \textbf{84.83}  & \\ 
			\specialrule{0em}{3pt}{0pt}
			\bottomrule
		\end{tabular}
	}
\end{table*}
\linespread{1.0}

\section{Experiments}

\subsection{Experimental Design}

This work primarily involves experiments with three types of models: 
a vanilla BERT with general pre-training, a MoE BERT with general pre-training, and a MoE BERT enhanced through transfer capability distillation. 
Among these, the vanilla BERT acts as a transfer capability teacher and also serves as a baseline model. 
The general pre-trained MoE BERT is the subject of our improvement and also a baseline model. 
The MoE BERT enhanced through transfer capability distillation is the model representing our method. 
We confirm the existence of transfer capability distillation and its effectiveness in improving the downstream task performance of MoE models by comparing the new model with two baseline models. 

We pre-trained two different sizes of BERT architectures, with the smaller size having 12 layers and a hidden dimension of 128, and the larger size having 12 layers and a hidden dimension of 768. 
We conducted experiments at both scales to ensure a more comprehensive validation. 
For both sizes, the number of experts in MoE was set to 64, and each hidden representation is processed only by the top-1 expert. 
For the larger model, we utilized all distillation losses, but for the smaller model, we did not use multi-head attention distillation loss (setting $\lambda_A$ to 0). 
This decision was based on our experimental observations, as we found it harmed transfer capability in the smaller scale. 

Our main experiments involved fine-tuning on downstream tasks using the GLUE benchmark, reporting results on the validation set. 
To address the potential issue of severe overfitting when fine-tuning MoE models directly, we performed both full-parameter fine-tuning and efficient adapter fine-tuning~\cite{pmlr-v97-houlsby19a} on all models, reporting the better result of the two for each model. 

More details on the pre-training and fine-tuning procedures can be found in Appendix \ref{appendix_a} and \ref{appendix_b}.

\subsection{Main Results}

For smaller-scale models (H=128), we enabled the vanilla BERT to undergo 20 epochs of pre-training and then used it as a transfer capability teacher to distill the MoE BERT for 5 epochs. 
For larger-scale models (H=768), we pre-trained the vanilla BERT for 10 epochs and then used it to distill the MoE BERT for 10 epochs. 

Regarding the MoE BERT with general pre-training, we pre-trained two models with different pre-training epochs for each scale, corresponding to two different settings: Pre-training Performance Alignment and Pre-training Epoch Alignment. 

\subsubsection{Pre-training Performance Alignment}

The first setting involves aligning the pre-training performance between a general pre-trained MoE BERT and a MoE BERT that has undergone transfer capability distillation. 
This is achieved by ensuring both models exhibit equivalent performance on the validation set of masked language modeling task, followed by comparing their downstream task performance. 
This setting allows for a more intuitive assessment of the improvement in the new model's transfer capability since their pre-training performances are identical. 

For smaller-scale models (H=128), the original MoE BERT was pre-trained for 6 epochs. 
For larger-scale models (H=768), the original MoE BERT was pre-trained for 12 epochs. 
The specific results are shown in Table \ref{table_1}. 

From these results, it's clear that for both sizes of models, 
the new models demonstrated significant improvements across all downstream tasks, confirming that our method indeed enhanced the transfer capability of MoE BERT. 
Notably, the MoE BERT with transfer capability distillation outperformed its teacher model in both pre-training and downstream task performance, indicating the existence of transfer capability distillation and validating our proposition that vanilla transformers are effective transfer capability teachers.

\linespread{1.6}
\begin{table}[t]
	\linespread{1.0}
	\caption{ The results of ablation analysis. \textbf{T}: Model Trunk, \textbf{I}: Residual Inner, \textbf{A}: Multi-Head Attention. }
	\linespread{1.6}
	\label{table_3}
	\resizebox{0.48\textwidth}{!}{
		\begin{tabular}{@{}clcccccccc@{}}
			\toprule
			& \textbf{Model} & \multicolumn{2}{c}{\textbf{MRPC}} & \multicolumn{2}{c}{\textbf{STS-B}}  & \multicolumn{2}{c}{\textbf{MNLI}} & \textbf{QNLI} & \\ 
			\midrule
			\multicolumn{9}{l}{\textit{(\textbf{H=128, \ Pre-Training Performance$\bm \approx$ –2.56})}} & \\
			& MoE BERT                        & 79.75 & 85.75 & 81.83 & 81.56 & 71.33 & 72.82 & 83.53  & \\ 
			& MoE BERT + T                    & 80.33 & 86.05 & 82.53 & 82.21 & 74.77 & 75.72 & 84.61  & \\ 
			& MoE BERT + T, I                 & \textbf{83.70} & \textbf{88.42} & \textbf{83.07} & \textbf{82.80} & \textbf{76.37} & \textbf{77.14} & \textbf{85.49}  & \\ 
			& MoE BERT + T, I, A \ \      & 82.87 & 88.02 & 82.96 & 82.74 & 75.45 & 76.25 & 85.40  & \\ 
			\midrule
			\multicolumn{9}{l}{\textit{(\textbf{H=768, \ Pre-Training Performance$\bm \approx$ –1.42})}} & \\
			& MoE BERT + T, I                 & 86.60 & 90.53 & 86.78 & 86.57 & 83.16 & 83.46 & 90.33 & \\ 
			& MoE BERT + T, I, A              & \textbf{87.58} & \textbf{91.28} & \textbf{87.38} & \textbf{87.10} & \textbf{83.48} & \textbf{83.70} & \textbf{90.80} & \\ 
			\bottomrule
		\end{tabular}
	}
\end{table}
\linespread{1.0}

\subsubsection{Pre-training Epoch Alignment}

The second setting involves aligning the actual pre-training epochs between a general pre-trained MoE BERT and a MoE BERT that has undergone transfer capability distillation. 
Since our new model requires pre-training a vanilla BERT teacher model before distillation, it effectively undergoes a greater amount of pre-training. 
Therefore, to validate the practical value of our method, we increased the pre-training epochs of the baseline MoE BERT to match the total of both the teacher and student model's pre-training epochs. 

For smaller-scale models (H=128), we increased the pre-training epoch of the original MoE BERT from 6 to 25. 
For larger-scale models (H=768), we increased it from 12 to 20. 
The specific results are shown in Table \ref{table_2}. 

For both sizes, the baseline MoE BERT, after more pre-training epochs, outperformed our new model in terms of pre-training performance. 
However, our model still significantly surpassed it in most downstream tasks. 
This not only further demonstrates that our method indeed enhances the transfer capability of MoE BERT, as it achieves stronger performance in downstream tasks despite weaker pre-training performance, 
but also confirms the practical value of our method under a more equitable setting of pre-training steps. 

\begin{figure}[!t]
	\centering
	\includegraphics[scale=0.45]{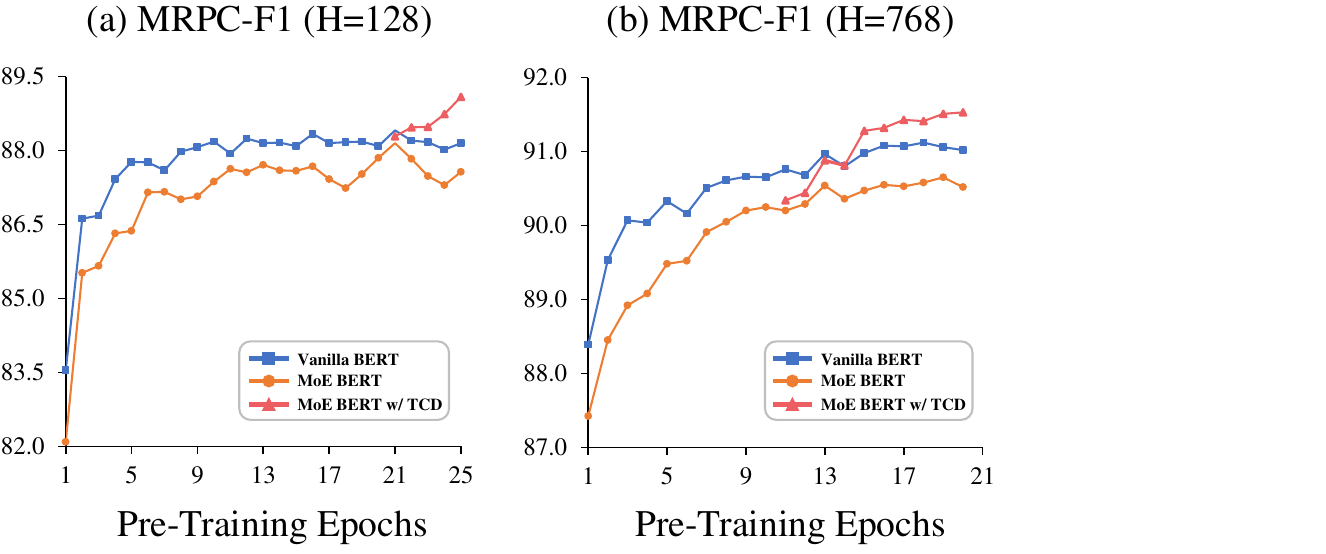}
	\caption{ Analysis of downstream performance trend. }
	\label{figure_5}
\end{figure}

\subsection{Ablation Analysis}

In our method, we select three locations for relation alignment: model trunk (T), residual inner (I) and multi-head attention (A). 
Here, we explore the necessity of constraints at these locations. 

For smaller-scale models (H=128), we incrementally added constraints to these three locations on the baseline MoE BERT. 
For larger-scale models (H=768), we compared the differences between adding and not adding multi-head attention location constraints. 
The performance comparison for all downstream tasks of the models is based on the aligned pre-training performance, which is also for intuitively reflecting the differences in transfer capability. 
The results are shown in Table \ref{table_3}. 

From the results in Table \ref{table_3}, we can see that constraints at the model trunk and residual inner are extremely important, leading to significant improvements in transfer capability. 
For smaller-scale models, the constraint at the multi-head attention location had a negative impact, so we ultimately did not use it in smaller-scale models. 
However, for larger-scale models, the constraint at the multi-head attention location showed a clear positive gain, so we implemented it in larger-scale models. 
But the general principles governing the effectiveness of multi-head attention location constraints are not yet fully clear, and we consider exploring this further in subsequent work. 

\subsection{Trend Analysis}

To more intuitively demonstrate the issue we are concerned with and the effectiveness of our method, we present the performance trend of various models on the MRPC task over increasing pre-training epochs, as specifically shown in Figure \ref{figure_5}. 

Firstly, we can clearly see that, whether in smaller or larger models, the baseline MoE BERT consistently underperforms the vanilla BERT on the MRPC task. This indicates a significant degradation in the transfer capability of MoE BERT, an issue that is the primary focus of this work. 

Then, MoE BERT, after undergoing transfer capability distillation, consistently outperforms the baseline MoE BERT on the MRPC task. This suggests that our method effectively enhances the transfer capability of MoE BERT and improves its downstream task performance. 

Finally, the performance of MoE BERT with transfer capability distillation, even surpasses that of the teacher model on the MRPC task. This validates our proposed idea of transfer capability distillation and proves that vanilla transformers are suitable transfer capability teachers.

\section{Transfer Capability Distillation vs. General Knowledge Distillation}
\label{section_compare}

Transfer capability distillation is evidently distinct from general knowledge distillation. 

\textbf{General knowledge distillation is a compression method}. It usually involves distilling from a larger model with either superior pre-training performance or stronger downstream performance, 
to create a smaller model that is relatively weaker in most aspects but more efficient. 

In this work, both the pre-training performance and downstream performance of vanilla models are weaker, and even the scales of vanilla models are smaller; 
they merely possess stronger inherent transfer capability. 
\textbf{We believe that small vanilla models can serve as transfer capability teachers, guiding distillation for larger MoE models with poorer transfer capability}. 
A distinctive characteristic of this approach is the counterintuitive outcome where a teacher model, inferior in pre-training performance and downstream performance, anomalously distills a student model superior in those aspects.
Therefore, fundamentally, 
\textbf{transfer capability distillation is not a compression method, but an enhancement method. }

\section{Why Does Transfer Capability Distillation Work?}

Although we propose transfer capability distillation and designed a distillation scheme that enhanced the transfer capability of MoE BERT, our understanding of the fundamental differences in transfer capability is quite limited. 
It is even difficult to explain why transfer capability can be distilled, which is clearly not conducive to further research. 

Here, we propose an explanation: the difference in transfer capability may be related to the quality of features learned during the pre-training phase of models, and \textbf{transfer capability distillation to some extent aligns student models' features with those high-quality features of teacher models}. 

Our viewpoint stems from the observation that the original MoE BERT, even without fine-tuning for downstream tasks and merely freezing parameters for masked language modeling task, exhibits significant differences from vanilla BERT. 

Specifically, we tested the models' masked language modeling capability on additional out-of-distribution (OOD) corpus, using the validation set of Pile dataset~\cite{gao2020pile}, which includes a wide range of corpus with significant distribution differences from the pre-training corpus, such as mathematics, GitHub, etc. 
The experiments were conducted on both scale models, ensuring alignment of pre-training performance before comparison, as shown in Table \ref{table_4}. 

It is not difficult to notice the out-of-distribution masked language modeling capability of original MoE BERT is significantly lower compared to vanilla BERT, whereas MoE BERT, after undergoing transfer capability distillation, shows a marked improvement in this regard. 
These results suggest that \textbf{even though models perform the same pre-training tasks, the quality of the learned features varies, which is likely the cause of differences in transfer capability}. 

Therefore, it is not difficult to understand why the distillation method is effective: 
it likely works by imposing additional constraints on the features, prompting MoE BERT to utilize higher-quality features for completing the pre-training tasks, which indirectly enhances its transfer capability.

\section{Related Work}

Our work is related to Mixture of Experts (MoE) models and general knowledge distillation. 

The MoE model is a type of dynamic neural network that excels in expanding model capacity with low computation cost. 
\citet{shazeer2017} added an MoE layer to LSTM, showing for the first time that MoE architecture can be adapted to deep neural networks. 
\citet{lepikhin2021gshard} enhanced machine translation performance using a Transformer model with the MoE architecture. 
\citet{JMLR_v23_21_0998} introduced the well-known Switch Transformers, demonstrating the application of MoE Transformers in pre-trained language models. 
\citet{artetxe-etal-2022-efficient} conducted extensive experiments on MoE Transformer, establishing its significant efficiency advantages over dense language model. 
Our work builds upon the existing MoE layer design, enhancing transfer capability in a non-invasive manner. 

General knowledge distillation primarily aims at reducing model size and computation costs. 
\citet{hinton2015distilling} initially proposed knowledge distillation, transferring knowledge learned on a large model to a smaller model. 
This concept was later adapted to pre-trained language models. 
\citet{sun-etal-2019-patient} compressed BERT into a shallower model through output distillation and hidden representation distillation. 
\citet{sanh2020distilbert} successfully halved the number of BERT layers through distillation during both pre-training and fine-tuning stages. 
\citet{jiao-etal-2020-tinybert} designed a distillation for BERT with multi-position constraints, also covering both stages. 
\citet{sun-etal-2020-mobilebert} proposed a method that retains transfer capability, offering greater versatility. 
Our work is different from general knowledge distillation, and the explanation is in Section~\ref{section_compare}.

\linespread{1.7}
\begin{table}[t]
	\linespread{1.0}
	\caption{ Masked language modeling results on out-of-distribution corpus. \textbf{AX}: ArXiv, \textbf{DM}: DM Mathematics, \textbf{GH}: Github, \textbf{SE}: Stack Exchange, \textbf{UI}: Ubuntu IRC. }
	\linespread{1.7}
	\label{table_4}
	\resizebox{0.48\textwidth}{!}{
		\begin{tabular}{@{}clcccccc@{}}
			\toprule
			& \textbf{Model}   & \textbf{AX} & \textbf{DM}  & \textbf{GH} & \textbf{SE} & \textbf{UI} & \\ 
			\midrule
			\multicolumn{7}{l}{\textit{(\textbf{H=128, \ Pre-Training Performance$\bm \approx$ –2.76})}} & \\
			& Vanilla BERT  & -3.545 & -2.955 & -3.462 & -3.530 & -4.120  & \\ 
			\cline{3-7}
			& MoE BERT                 & -3.613 & -3.026 & -3.564 & -3.585 & -4.164  & \\ 
			& MoE BERT w/ TCD \ \ \ \  & \ \textbf{-3.563} \ & \ \textbf{-2.959} \ & \ \textbf{-3.499} \ & \ \textbf{-3.536} \ & \ \textbf{-4.118} \  & \\ 
			\midrule
			\multicolumn{7}{l}{\textit{(\textbf{H=768, \ Pre-Training Performance$\bm \approx$ –1.57})}} & \\
			& Vanilla BERT  & -2.338 & -2.179 & -2.420 & -2.443 & -3.052 & \\ 
			\cline{3-7}
			& MoE BERT                 & -2.393 & -2.296 & -2.481 & -2.481 & -3.121 & \\ 
			& MoE BERT w/ TCD          & \textbf{-2.334} & \textbf{-2.219} & \textbf{-2.426} & \textbf{-2.444} & \textbf{-3.051} & \\ 
			\bottomrule
		\end{tabular}
	}
\end{table}
\linespread{1.0}

\section{Conclusion}

This work focuses on the issue of MoE transformers underperforming in downstream tasks compared to vanilla transformers. 
We propose that the model's pre-training performance and transfer capability are different factors affecting downstream task performance, and the root cause of the MoE model's poor performance in downstream tasks is its inferior transfer capability. 
To address it, we introduce transfer capability distillation, utilizing vanilla models as teachers to enhance the transfer capability of MoE models. 
We design a distillation scheme that solves the issue of weak transfer capability in MoE models, improving performance in downstream tasks and confirming the concept of transfer capability distillation. 
Finally, we provide insights from model feature perspective to explain our method, offering ideas for future research.

\section{Limitations}

Although this work introduces the concept of transfer capability distillation and addresses the issue of weak transfer capability in MoE Transformers, there are still some limitations. 

1. We pre-trained the teacher model to a level we consider appropriate and demonstrated the feasibility of transfer capability distillation through experiments. 
However, the level of pre-training of the teacher model may affect the effect of transfer capability distillation, and uncovering this pattern could be very helpful for practical applications. 
We plan to explore this in our future work. 

2. While we have conducted experiments on models of two different sizes and carried out rigorous validation, due to limited resource, we have not pre-trained or tested the models with more parameters. 
We consider addressing this issue through future collaborations. 

3. Although we have hypothesized about the reasons why transfer capability distillation works, more evidence is needed. 
We plan to delve into this issue in our subsequent research.

\bibliography{anthology,custom}

\appendix

\newpage

\section{Pre-training Procedure}
\label{appendix_a}

All experiments were conducted in English only. 
This work utilized the same pre-training corpus as that of~\cite{devlin-etal-2019-bert}, namely Wikipedia and BooksCorpus~\cite{Zhu_2015_ICCV}. 
A subset of this pre-training corpus was randomly selected as a validation set to represent the performance of models during pre-training. 

For the masked language modeling task, we adopted the same approach as~\cite{devlin-etal-2019-bert}. 
Specifically, 15\% of the tokens in a sequence were selected for masking, with 80\% of these replaced by the [MASK] token, 10\% substituted with random tokens, and the remaining 10\% left unchanged. 
Differing from the method proposed by~\cite{devlin-etal-2019-bert}, we omitted the next sentence prediction task, and instead used longer continuous text segments as our pre-training input sequences. 
Additionally, different masking schemes were applied to the same input sequence in different epochs. 

Our smaller-scale models have a hidden dimension of 128, 12 layers, 2 attention heads and 6.3M Parameters. 
Our larger-scale models have a hidden dimension of 768, 12 layers, 12 attention heads and 110M Parameters. 
The maximum sequence length for all models is 128 tokens. 
All models use the same vocabulary as the BERT model published by~\cite{devlin-etal-2019-bert}, containing 30,522 tokens. 
Each of our MoE models has 64 experts. 
The smaller-scale (H=128) MoE models have 105M Parameters, and the larger-scale (H=768) MoE models have 3.6B Parameters. 
We employed the FastMoE framework proposed by~\cite{he2021fastmoe,he2022fastermoe} for the implementation of MoE BERT models. 
In addition, we also used the PyTorch\footnote{https://pytorch.org/} and transformers\footnote{https://github.com/huggingface/transformers} libraries. 

For all MoE BERT models, \(\lambda_B\) was set to 1000. 
For MoE BERT models undergoing transfer capability distillation, \(\lambda_T\) and \(\lambda_I\) were both set to 1; for larger-scale models, \(\lambda_A\) was set to 1, while for smaller-scale models, \(\lambda_A\) was set to 0. 
The relation constraints at the model trunk and residual inner required sampling tokens from a batch. 
We sampled 4096 tokens, divided into 32 groups, with each group comprising 128 representations for pairwise cosine similarity calculations. 

All models were pre-trained for a maximum of 40 epochs, although this maximum was not reached in practice. 
Some checkpoints from specific epochs were selected for alignment and experimentation. 
Pre-training for all models was conducted using the Adam optimizer~\cite{DBLP_journals_corr_KingmaB14}, with a learning rate of \(1 \times 10^{-4}\), \(\beta_1 = 0.9\), \(\beta_2 = 0.999\), an L2 weight of 0.01. 
The learning rate warmed up in the first 10,000 steps, followed by linear decay. 
The smaller-scale models were pre-trained with a batch size of 512 on 4 x Nvidia Tesla V100 GPUs, and total GPU days are approximately 42 days. 
The larger-scale models were pre-trained with a batch size of 1024 on 4 x Nvidia Tesla A100 GPUs, and total GPU days are approximately 98 days. 

To ensure a fair comparison, all models were pre-trained from scratch. 
However, due to limited computational resources, our pre-training tokens were generally fewer than those in the original BERT paper~\cite{devlin-etal-2019-bert}, which may lead to some discrepancies in the downstream task results compared to the original BERT paper.

\section{Fine-tuning Procedure}
\label{appendix_b}

We conducted fine-tuning experiments on the GLUE benchmark~\cite{wang-etal-2018-glue}. 
The maximum number of training epochs for all models was set to 10, with a batch size of 32. 
The optimizer was Adam~\cite{DBLP_journals_corr_KingmaB14}, with a warmup ratio of 0.06, a linearly decaying learning rate, and a weight decay of 0.01. 
We reported the average performance of multiple runs. 

For full parameter fine-tuning, the learning rates were \{1e-5, 2e-5, 5e-5\}. 
For adapter fine-tuning, the learning rates were \{1e-4, 2e-4, 3e-4\}. 
The adapter sizes for the small models (H=128) were \{16, 64, 128\}, and for the large models (H=768) were \{16, 64, 128, 256\}. 

Additionally, there were some exceptions. 
For the MNLI, QNLI, and QQP datasets, a small number of fine-tuning epochs in small models during adapter fine-tuning limited performance, so we increased the maximum training epochs to 20. 
For the MNLI dataset, using a small adapter size in small models during adapter fine-tuning limited performance on MNLI, so we included an experiment with an adapter size of 512. 

\end{document}